\documentclass[conference]{IEEEtran}
\pdfoutput=1
\usepackage[pdftex]{graphicx}
\usepackage{algpseudocode}
\usepackage{algorithmicx}
\usepackage{amsmath,mathtools}
\usepackage{flushend}
\usepackage{float}

\begin{document}

\title{What Does a TextCNN Learn?}

\author{
	\IEEEauthorblockN{Gong, Linyuan}
	\IEEEauthorblockA{
		Peking University
	}
	\and
	\IEEEauthorblockN{Ji, Ruyi}
	\IEEEauthorblockA{
		Peking University
	}
}

\maketitle



\section{Introduction}
TextCNN, the convolutional neural network for text, is a useful deep learning algorithm for sentence classification tasks such as sentiment analysis and question classification\cite{Kim2014ConvolutionalNN}.

However, neural networks have long been known as “black boxes” because interpreting them is a challenging task. Researchers have developed several tools to understand a CNN for image classification by deep visualization\cite{Yosinski2015UnderstandingNN}, but research about deep TextCNNs is still insufficient.

In this paper, we are trying to understand what a TextCNN learns on two classical NLP datasets. Our work focuses on functions of different convolutional kernels and correlations between convolutional kernels.

\section{Model and Training Configurations}

Here is the structure of our two-layer TextCNN model:

\begin{figure}[htbp]
	\centering
	\begin{tabular}{|c|c|c|}
	\hline
	\multicolumn{3}{|c|}{softmax}  \\ \hline
	\multicolumn{3}{|c|}{dropout}  \\ \hline
	max-pool & max-pool & max-pool \\ \hline
	relu     & relu     & relu     \\ \hline
	bn       & bn       & bn       \\ \hline
	conv (3) & conv (4) & conv (5) \\ \hline
	relu     & relu     & relu     \\ \hline
	bn       & bn       & bn       \\ \hline
	conv (3) & conv (4) & conv (5) \\ \hline
	\multicolumn{3}{|c|}{embed}    \\ \hline
	\end{tabular}
	\caption{The network structure of our TextCNN model; conv: convolutional layers with filter window $n$ and $64$ feature maps each; bn: batch normalization; relu: rectified linear units; max-pool: turns a 2-D matrix into a 1-D vector keeping maximum values of each column; dropout: keep rate is $0.5$}
\end{figure}

The embed layer translates a word into a 300-dimension vector. Word vectors are pre-trained on GoogleNews by Word2Vec\cite{website:word2vec}. The embed layer is \textit{static}: word vectors does not change during training.

For all datasets, we use stochastic gradient descent with a mini-batch size of $128$ and the Adam update rule\cite{Kingma2014AdamAM}. We trained for $10$ epochs on all training data. We did not do any dataset-specific hyperparameter tuning.

\section{Dataset and Training Results}

We trained our model on two datasets:

\begin{LaTeXdescription}
	\item[TREC] classify a question into 6 question types (whether the question is about persons, location, numeric information, etc.)\cite{Li2002LearningQC}.
	\item[SST] Stanford Sentiment Treebank. Classify a movie review with one sentence as positive, normal, or negative (we relabeled the dataset)\cite{SocherEtAl2013:RNTN}.
\end{LaTeXdescription}

Here are some statistics about datasets and our training results:

\begin{figure}[htbp]
	\centering
	\begin{tabular}{l|l|l|l|l}
	\hline
	\textbf{Data} & $|C|$ & $|D_{train}|$ & $|D_{test}|$ & \textit{Acc} \\ \hline
	\textbf{TREC} & 6     & 5452          & 500          & 0.924        \\ 
	\textbf{SST}  & 3     & 9076          & 2210         & 0.6027       \\ \hline
	\end{tabular}
	\caption{Summary statistics of datasets and training results. $|C|$: the number of classes. $|D_{train}|$: the number of training data. $|D_{test}|$: the number of test data. \textit{Acc}: top-1 accuracy.}
\end{figure}

\section{Our Method}
We extract all n-grams in the dataset (skipping words which are not in the GoogleNews dictionary) and feed them into each convolutional kernel with the appropriate filter window ($h$-grams for the first layer and $(2h-1)$-grams for the second layer). We record activation values of each convolutional kernel (after bn and relu layers) and do some statistical analysis.

\section{Comparing Convolutional Kernels}
\subsection{Label the Kernels}
For each kernel, we find out n-grams which generate top-3 activation values. If they come from the sentences with the same label, then we classify this kernel as this label. If the top-3 n-grams come from a mixed set of sentences, then we classify this kernel as the type ``other.'' Here are experiment results of all 6 sets of convolutional kernels of TREC dataset:

\begin{figure}[htbp]
	\begin{tabular}{l|llll|llll}
		\hline
		\textbf{Class} & \textbf{1-3} & \textbf{1-4} & \textbf{1-5} & \textbf{Sum} & \textbf{2-3} & \textbf{2-4} & \textbf{2-5} & \textbf{Sum} \\ \hline
		\textbf{ABBR}  & 2            & 6            & 1            & 9            & 0            & 12           & 0            & 12           \\
		\textbf{DESC}  & 4            & 5            & 7            & 16           & 6            & 14           & 6            & 26           \\
		\textbf{ENTY}  & 6            & 5            & 6            & 17           & 5            & 2            & 7            & 14           \\
		\textbf{HUM}   & 16           & 2            & 6            & 24           & 19           & 0            & 13           & 32           \\
		\textbf{LOC}   & 7            & 3            & 2            & 12           & 9            & 9            & 3            & 21           \\
		\textbf{NUM}   & 7            & 6            & 6            & 19           & 13           & 9            & 14           & 36           \\
		\textbf{Other} & 22           & 37           & 36           & 95           & 12           & 18           & 21           & 51           \\ \hline
	\end{tabular}
	\caption{The number of convolutional kernels of each class in each layer. 1-3 stands for the first layer of the tower with filter window $3$.}
\end{figure}

This table tell us:

\begin{itemize}
	\item More than half of the kernels have a preference for one specific label. It means that kernels have learned division of labor.
	\item The number of kernels in different classes varies. This variation reflects subtle characteristics of each class. For example:
	\begin{itemize}
		\item Questions about humans(HUM) and numbers(NUM) are relatively harder to recognize than questions about abbreviations(ABBR) are, because the former two classes require far more kernels than the latter one does.
		\item Features about abbreviations(ABBR) are usually associated with 4-grams because almost all kernels of type ABBR in the second layer are in the tower with filter window $4$.
	\end{itemize}
	\item There are more kernels in the ``other'' class in the first layer than in the second layer. It implies that more kernels in the first layer learn some generic features, while more kernels in the second layer learn label-specific features.
\end{itemize}

\subsection{Generic Features}
Kernels in the ``other'' class learn some generic features that are shared by multiple labels. To gain some intuition about these kernels, we observed top-3 sensitive n-grams of kernels in both models(trained on TREC and SST) and found out some typical examples:

\begin{figure}[htbp]
	\centering
	\begin{tabular}{lll}
	\hline
	\multicolumn{3}{l}{\textbf{SST/1-3/\#36}} \\ \hline
	an     & invigorating     & electric      \\
	is     & refreshing       & absence       \\
	an     & unending         & soundtrack    \\ \hline
	\end{tabular}

	\vspace{0.05in}

	\begin{tabular}{llll}
	\hline
	\multicolumn{4}{l}{\textbf{SST/1-4/\#3}}     \\ \hline
	explores & the  & difficult   & relationship \\
	defuses  & this & provocative & theme        \\
	reducing & our  & emotional   & stake        \\ \hline
	\end{tabular}
	\caption{Two ``other'' kernels that recognizes grammar patterns. The first one prefers is/an + adj. + n. and the second one prefers v. + art. + adj. + n.}
\end{figure}

\begin{figure}[htbp]
	\centering
	\begin{tabular}{llll}
	\hline
	\multicolumn{4}{l}{\textbf{SST/1-4/\#38}}     \\ \hline
	nonstop & romance     & music      & suspense \\
	this    & intricately & structured & drama    \\
	the     & best        & sex        & comedy   \\ \hline
	\end{tabular}

	\vspace{0.05in}

	\begin{tabular}{lllll}
	\hline
	\multicolumn{5}{l}{\textbf{TREC/1-5/\#23}} \\ \hline
	Doodle  & why  & did   & Yankee  & Doodle  \\
	in      & the  & song  & Yankee  & Doodle  \\
	make    & up   & half  & the     & Soviet  \\ \hline
	\end{tabular}
	\caption{Two ``other'' kernels that recognizes topics. The first one prefers various types of drama; the second one prefers concepts closely related to a nation at the place of the last two words.}
\end{figure}

Some ``other'' kernels focus on overall grammar patterns of n-grams, while others focus on general topics related to n-grams. However, the boundary between these two types are vague, and there are still many ``other'' kernels whose functions remain unknown to humans.

\section{Correlations Between Kernels}
\subsection{The Measurement of Correlations}
Suppose the activation value of the i-th kernel filled with the k-th n-gram is $a_{ik}$. We compute the correlation coefficient $r$ between vector $(a_{i1},a_{i2},\dots)$ and vector $(a_{j1},a_{j2},\dots)$. The larger $|r|$ is, the closer the relationship between these two kernels is. Because every $r$ is either positive or nearly zero, we omit the symbols for absolute value.

We can also visualize the correlation between two kernels as an activation graph:

\begin{enumerate}
	\item Sort each pair of activation $(a_{i1}, a_{j1}), (a_{i2}, a_{j2}), \dots$ by the first keyword and keep the first 1200 pairs.
	\item Split 1200 pairs into 3 slices and sort each slice by the second keyword.
	\item Plot the values of each pair.
\end{enumerate}

Here is an example:

\begin{figure}[htbp]
	\includegraphics[width=\hsize]{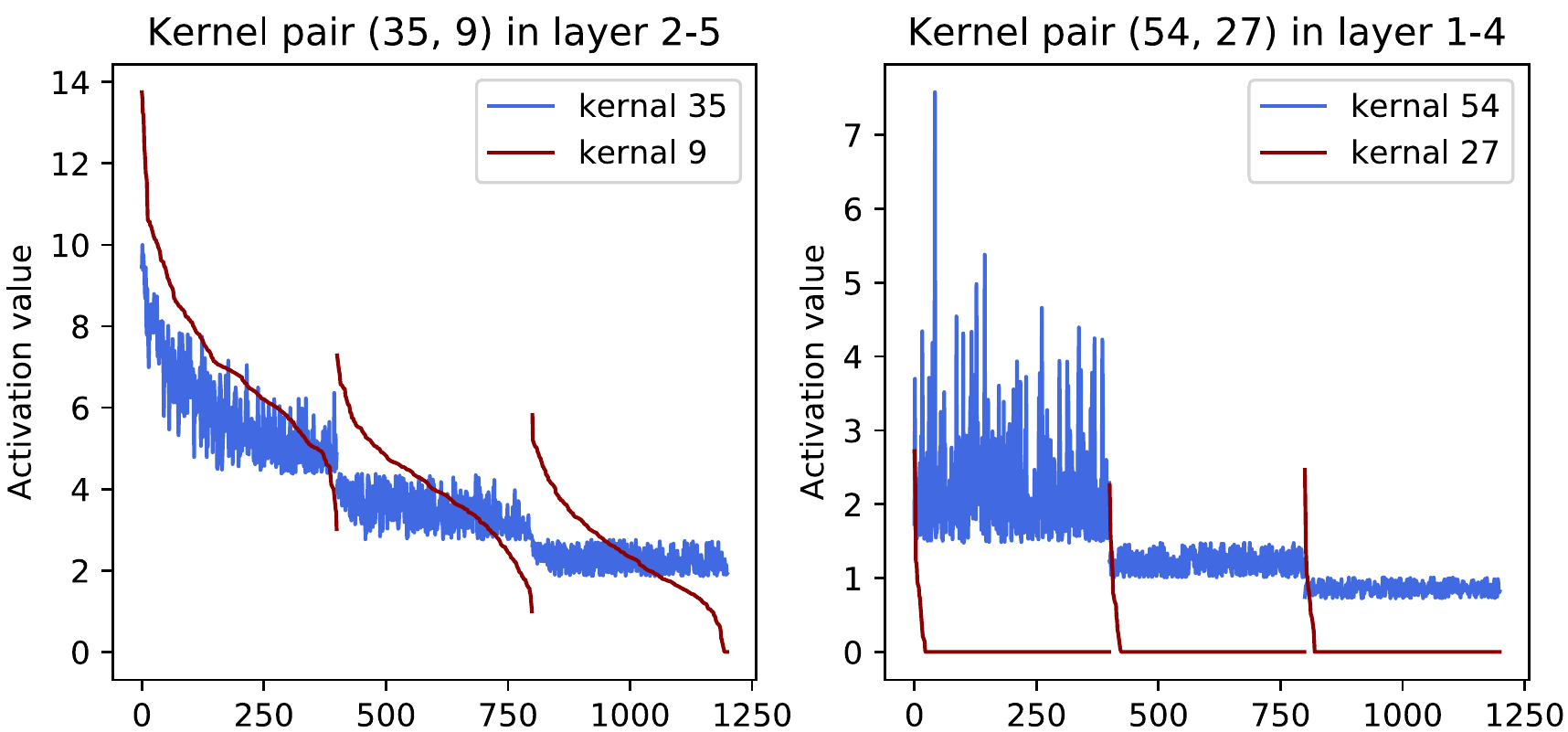}
	\caption{The activation graphs of two pairs of kernels. For the first pair, $r=0.94404$, and their lines are close to each other; for the second pair, $r=0.00002$, and their lines are far from each other.}
\end{figure}

We found that the first pair of kernels recognizes similar n-grams while the second pair of kernels recognizes completely different n-grams. This example illustrates that our measurements can reflect correlations between kernels.

\subsection{Correlated Pairs}
After computing correlations between all pairs of kernels in our model for TREC, we get the following results:

\begin{figure}[htbp]
	\centering
	\begin{tabular}{l|llll|llll}
	\hline
	$r$     & \textbf{1-3} & \textbf{1-4} & \textbf{1-5} & \textbf{Sum} & \textbf{2-3} & \textbf{2-4} & \textbf{2-5} & \textbf{Sum} \\ \hline
	$> 0.8$ & 11           & 7            & 21           & 39           & 312          & 92           & 215          & 619          \\
	$> 0.6$ & 34           & 41           & 90           & 165          & 610          & 313          & 331          & 1254         \\ \hline
	\end{tabular}
	\caption{The number of pairs of kernels in each layer whose $r$ is greater than $0.8$ or $0.6$.}
\end{figure}

Data implies that the correlations between kernels in the second layer are stronger than those in the first layer. If the discrepancy is due to redundancy, we may decrease the number of kernels in the second layer to compress the model. Therefore, our correlation analysis has the potential to help optimize networks.

\subsection{Correlated Tuples and ``Bridges''}
Kernel $k$ is a \textit{bridge} connecting kernel $i$ and kernel $j$ if $r(i,j)<0.1$ but $r(i,k)>0.4$ and $r(j,k)>0.4$. Here is an example of a bridge in the L1-3 in our model for TREC:

\begin{figure}[H]
	\centering
	\includegraphics[width=\hsize]{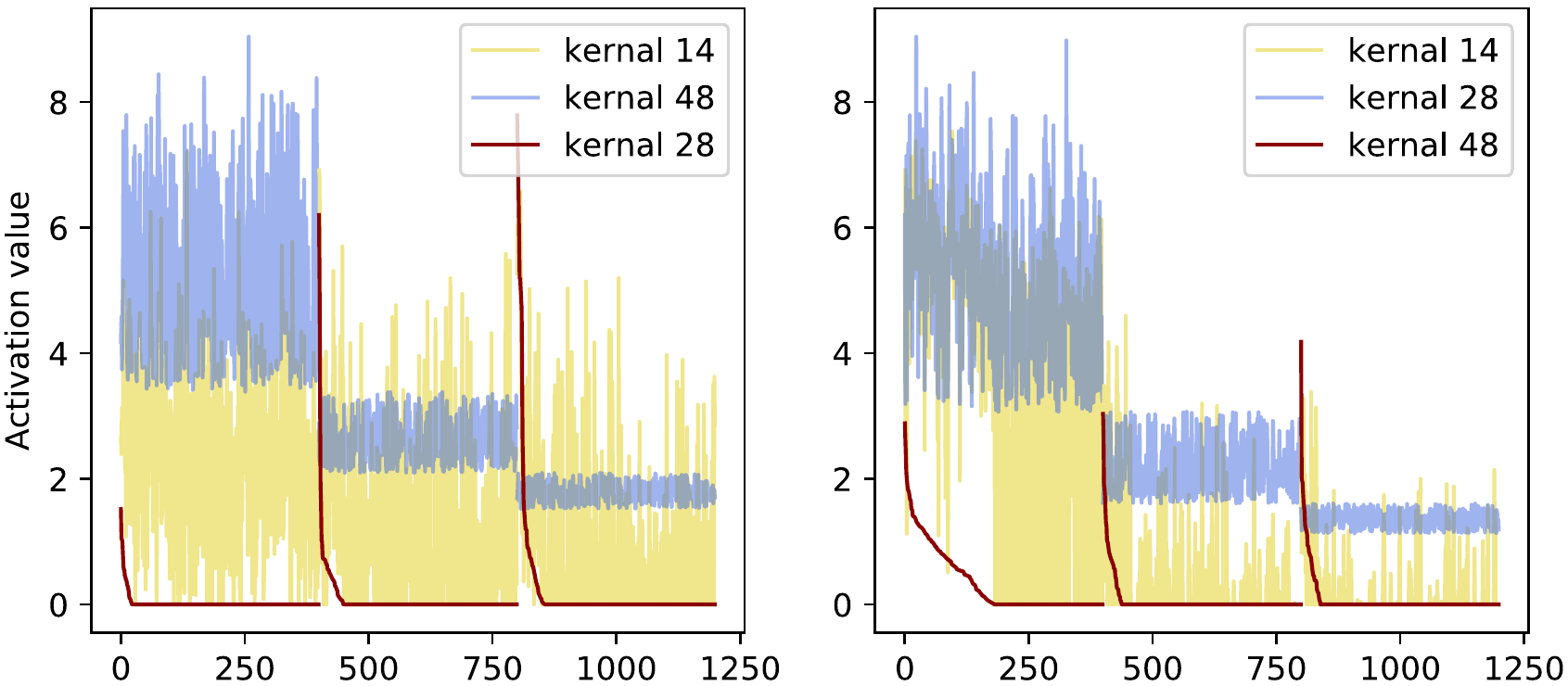}
	\caption{Kernel 14 is the bridge between kernel 28 and kernel 48 because $r(14,28)=0.5656$, $r(14,48)=0.4801$, and $r(28,48)=0.0012$.}
\end{figure}

The top-4 sensitive n-grams of these kernels also reflect this relationship:

\begin{figure}[htbp]
	\centering
	\begin{tabular}{lll}
	\hline
	\textbf{1-3/\#48}  & \textbf{1-3/\#14} & \textbf{1-3/\#28} \\ \hline
	\small{Who appointed the}  & \small{How many equal}    & \small{How many months}   \\
	\small{Who bestowed great} & \small{How many hostages} & \small{is average salary} \\
	\small{Who murdered Leno}  & \small{How many soldiers} & \small{How many seconds}  \\
	\small{Who penned Neither} & \small{Who lives at}      & \small{How many hours}    \\ \hline
	\end{tabular}
	\caption{Kernel 48 recognizes questions about humans and Kernel 28 recognizes questions about numbers. Kernel 14, the bridge between them, recognizes patterns related to both topics.}
\end{figure}

We counted the number of bridges in each layer in our model for TREC and found that the number of bridges in the second layer is more than the number of bridges in the first layer. It implies that the kernels in the second layer have a stronger tendency to cooperate, but it may also be a signal of redundancy.

\begin{figure}[htbp]
	\centering
	\begin{tabular}{l|llll}
	\hline
				& \textbf{3} & \textbf{4} & \textbf{5} & \textbf{Sum} \\ \hline
	\textbf{L1} & 6          & 6          & 30         & 42           \\ 
	\textbf{L2} & 1212       & 179        & 251        & 1642         \\ \hline
	\end{tabular}
	\caption{Number of bridges in each layer.}
\end{figure}

\section{Conclusion}
We trained a TextCNN for classifying texts and used some quantitative approaches to analyze the relationship between kernels. Our method is not restricted to convolution kernels, though, and it may help analyze other structures such as fully connected layers.

Using our method, we got some results about TextCNN: kernels learn features about labels; some kernels are analogous; some kernels learn common features of different classes; the depth of the layer influences the learned features.

\bibliographystyle{abbrv}
% edit reference.bib to add entries
\bibliography{reference}

\end{document}